\documentclass[letterpaper, 10 pt, conference]{ieeeconf}  
\IEEEoverridecommandlockouts                              

\usepackage{graphics}    
\usepackage{times}       
\usepackage{amsmath}     
\usepackage{amssymb}     
\usepackage{graphicx}
\usepackage{subcaption} 
\usepackage[separate-uncertainty=true,range-units=single]{siunitx}%
\usepackage{tabularx}
\usepackage{booktabs}
\usepackage{multirow}
\usepackage{bigdelim,bigstrut}
\usepackage{cite}

\usepackage[hidelinks]{hyperref} 

\hypersetup{
	colorlinks=true,        
	linkcolor=black,        
	citecolor=black,        
	filecolor=black,        
	urlcolor=black          
}

\usepackage{flushend}

\usepackage[font=small]{caption}

\def\eqref#1{Eq.~(\ref{#1})}

\renewcommand{\u}[1]{\text{ }\unit{#1}}
\newcommand\etal{\emph{et al. }}

\title{\LARGE \bf EnQuery: Ensemble Policies for Diverse Query-Generation \\ in Preference Alignment of Robot Navigation}

\author{Jorge de Heuvel \and  Florian Seiler \and Maren Bennewitz
  \thanks{All authors are with the Humanoid Robots Lab and the Center for Robotics at the University of Bonn, Germany. 
  	M. Bennewitz and J. de Heuvel are additionally
    with the Lamarr Institute for Machine Learning and Artificial
    Intelligence, Bonn, Germany. 
    This work has partially been funded by the Deutsche Forschungsgemeinschaft (DFG, German Research Foundation) under the grant number \mbox{BE~4420/2-2}~(FOR 2535 Anticipating Human Behavior) and by the Federal Ministry of Education and Research~(BMBF) under the grant number 16KIS1949.}
}

\begin{document}
\maketitle
\thispagestyle{empty} 
\pagestyle{empty}

\begin{abstract} 
To align mobile robot navigation policies with user preferences through reinforcement learning from human feedback (RLHF), reliable and behavior-diverse user queries are required. 
However, deterministic policies fail to generate a variety of navigation trajectory suggestions for a given navigation task. 
In this paper, we introduce EnQuery, a query generation approach using an ensemble of policies that achieve behavioral diversity through a regularization term. 
For a given navigation task, EnQuery produces multiple navigation trajectory suggestions, thereby optimizing the efficiency of preference data collection with fewer queries. 
Our methodology demonstrates superior performance in aligning navigation policies with user preferences in low-query regimes, offering enhanced policy convergence from sparse preference queries.
The evaluation is complemented with a novel explainability representation, capturing full scene navigation behavior of the mobile robot in a single plot.
Our code is available online at \href{https://github.com/hrl-bonn/EnQuery}{https://github.com/hrl-bonn/EnQuery}.
\end{abstract}

\section{Introduction}
\label{sec:intro}

For optimal human-robot interactions, robots should customize to the user needs.
Where policies demonstrating superior capabilities are developed through learning-based systems, the need for their preference-alignment methods arises~\cite{christiano_deep_2017, kaufmann_survey_2023}.
Reinforcement learning from human feedback (RLHF) is the state-of-the-art method, where user preferences are transferred into a reward model that aligns policies that interact with the human in the field of large language models~\cite{ziegler_fine-tuning_2020}, or robotics~\cite{cabi_scaling_2020, marta_variquery_2023, marta_sequel_2024, holk_polite_2024}.
In the context of mobile robot navigation, humans exhibit diverse preferences about comfortable robot approaching behavior and proxemics~\cite{syrdal_personalized_2007}, calling preference-aligned navigation policies on the plan~\cite{keselman_optimizing_2023, marta_aligning_2023, de_heuvel_learning_2023-1, de_heuvel_learning_2022}.

A core optimization goal in RLHF is to maximize information gain achieved by querying the user~\cite{kaufmann_survey_2023, biyik_asking_2020}.
Not only does this minimize the effort and fatigue associated with repetitive queries of the user, but enhances the quality of collected preference data.
Besides query diversity~\cite{biyik_batch_2018}, a reliable query test result is essential~\cite{biyik_asking_2020}.
However, inconsistent query results due to a low reliability undermine their corresponding preference information gain.
The reliability can however be increased when all test variables are kept constant across measurements.  
So to extract consistent user preferences, it is advisable to minimize changes in variables associated with a single query.
One approach is to keep the task environment constant while only altering the agent's behavior. This improves reliability and quality of information extracted from the user feedback.

\begin{figure}[t]
	\centering
	\includegraphics[width=1.0\linewidth]{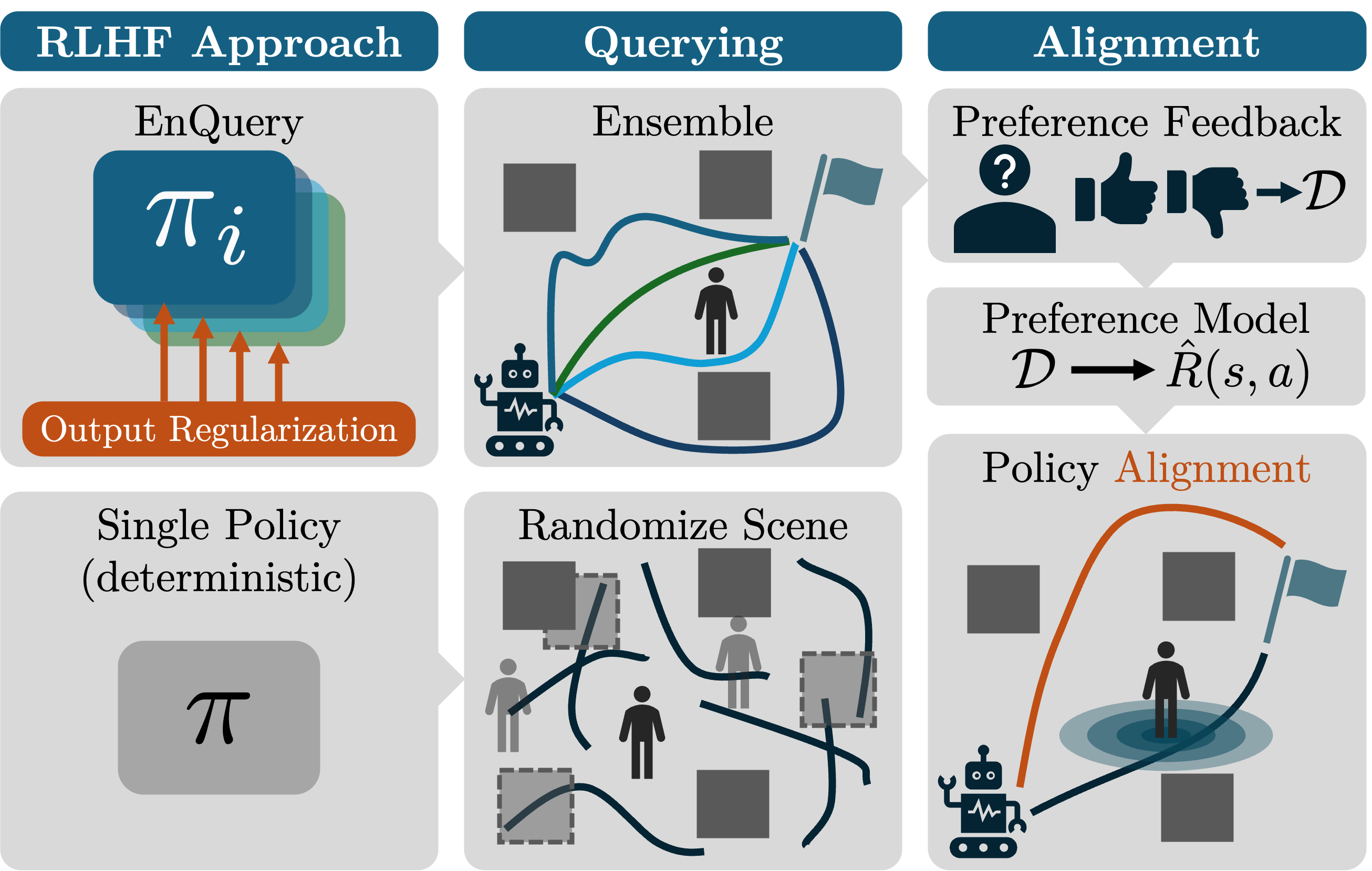}
	\caption{
		Our ensemble of RL policies generates a variety of trajectories for a given navigation task as queries for RL from human feedback.
		In contrast, deterministic policies are limited to just one trajectory, and the queries' variety depends on trajectory segments from randomized scene configurations.
		As a result, EnQuery facilitates a higher preference information gain for low query numbers.
	}
	\label{fig:motivation}
\end{figure}

In a typical robotics RLHF setup employing a deterministic policy, a pool of query trajectories is generated using policy rollouts of different environment configurations~\cite{christiano_deep_2017}.
Subsequently, the trajectories are subsampled into segments and presented to the user as pairwise preference queries. 
However, the diversity of environment task configurations in the query pool and the lack of common reference points in the segments conflict with the concept of re-test reliability through minimal change in variables, as elaborated above.

In contrast for the alignment process of large language models (LLM), it has become a best practice to ask for user ranking between two different outputs for a single input prompt~\cite{wang_aligning_2023}.
This is possible due to the generative and non-deterministic nature of these LLMs.
With deterministic policies however, this approach would result in two identical outputs, leaving no room for a preference choice of the user.

In this paper, we therefore present ensemble policies as possible solution for query diversity under identical policy-input on deterministic policies, see Fig.~\ref{fig:motivation}.
On a given navigation task, we train the ensemble with a regularization term that encourages dissimilar outputs of the individual ensemble members.

We apply the proposed method to a human-centric robot navigation task employing ensemble queries to capture distance-related preferences. 
We comprehensively test this approach starting with an analysis of the queries generated by the ensemble. 
We then assess a reward model trained on these queries and finally align a baseline-objective navigation policy to reflect the collected preferences.

The \textbf{main contributions} of our work are the following:
\begin{itemize}
	\item EnQuery, an ensemble approach for query generation enabled by a regularization term that ensures behavioral ensemble diversity.
	\item An extensive quantitative and qualitative analysis of the full query-to-alignment pipeline demonstrating superiority of our method in preference reflection for low query numbers, compared to a state-of-the-art baseline approach, ultimately enabling more query-efficient feedback processes.
	\item A novel visualization scheme for enhanced behavior explainability of mobile robot navigation policies of mobile robots that captures full scene behavior in a single plot.
\end{itemize}

\section{Related Work}
\label{sec:related}
In the domain of robotics, reinforcement learning from human feedback (RLHF) presents a promising solution to the challenges of defining and optimizing reward signals that maximize user preferences.
By incorporating human judgments, RLHF enables robots to adapt their learning objectives, ensuring behaviors align with human preferences. 
This method has shown significant potential in enhancing interaction tasks, robot manipulation~\cite{cabi_scaling_2020}, and robot navigation~\cite{keselman_optimizing_2023, marta_variquery_2023}.
Here, open challenges are the optimization of the human feedback process with respect to efficiency, information gain, and reduction of psychological biases~\cite{kaufmann_survey_2023}.

When it comes to the choice of queries in RLHF, typically a pool of randomly generated trajectories is used.
With the goal of optimal information gain in mind, query selection algorithms have emerged to surpass the naive approach of random sampling. 
Christiano \etal~\cite{christiano_deep_2017} choose queries either via uniform sampling or to maximize the variance in an ensemble of reward model.
Marta \etal~\cite{marta_variquery_2023} maximize the distance in a latent space representation of the trajectories generated through a variational autoencoder.
In contrast, we do not perform a selection of queries from a random pool based on certain criteria, but directly generate trajectory queries using the behavior-diverse policy ensemble for a give environment configuration.
Furthermore, psychological factors influence the preference and preference-consistency of the human.
As such, serial position effects can cause start and end of a trajectory to over-proportionally influence the user~\cite{kaufmann_survey_2023}. 
We counteract this effect with the design our navigation ensemble query approach with the alignment of start and end, where the trajectories originate from a common starting point and converge again at the goal location.

Typically, ensemble strategies are used for measuring uncertainty in learning-based models~\cite{gawlikowski_survey_2023}.
In the scope of reinforcement learning, applications of ensembles aim to improve the learning process, more specifically by stabilizing Q-learning and balancing exploration and exploitation~\cite{lee_sunrise_2021}. 
Lee \etal~\cite{lee_sunrise_2021} randomize the model weight initialization and bootstrap the data presented to each policy during an update.
Sheikh \etal~\cite{sheikh_maximizing_2021} encourage representation diversity through regularization terms with a similar goal.
The lack of such regularization terms was found to cause alignment of the ensemble members over the course of training, even if the networks are differently initialized.
In our work, we apply a regularization term on the policy outputs in the ensemble of TD3 agents for maximum behavioral diversity on a given input.

\section{Preliminaries}
\label{sec:approach}
\subsection{Problem Definition}
We consider a social robot navigation scenario, where the robot pursues a goal in a human environment among static unknown obstacles.
The robot is aware of the location of a single human in its vicinity and uses a 2D~lidar data to sense obstacles.
A user has specific preferences about the navigational behavior of the robot with respect to proxemics and taken paths and expresses them in pairwise comparisons of trajectory shapes.
We learn a behavior-diverse policy ensemble, in which each policy's linear and angular velocity commands take a different trajectory for a given navigation task with start and goal position, while avoiding collisions with obstacles and the human.
The resulting trajectory options represent trajectories for A$\succ$B preference comparisons.
The navigation policy is obtained using reinforcement learning, as elaborated  below.

\subsection{Reinforcement Learning of Point Navigation}
In reinforcement learning, the objective is to optimize state transitions $s_t \rightarrow s_{t+1}$ of a Markov Decision Process, leading to a reward $r_t$ for executing action $a_t = \pi(s_t)$ at time step~$t$, based on the policy $\pi$. 
These sequences $\left(s_t, a_t, r_t, s_{t+1}\right)$ are identified as state-action pairs. 
The optimization goal is to maximize the total return $R = \sum^{T}_{i=t} \gamma^{(i-t)}r_t$, which represents the sum of $\gamma$-discounted rewards from time $t$ onward. 
We use the Twin Delayed Deep Deterministic Policy Gradient (TD3) framework for continuous action spaces.
We employ the TD3 implementation of Stable-Baselines3~\cite{raffin_stable-baselines3_2021}.
All important parameters and notations of our work are listed in Table~\ref{tab:params}.

\subsubsection{State Space}
The state space includes data on the goal, the human in the vicinity of the robot, and nearby obstacles detected by the lidar sensor.
More specifically, our agents observe the environment through a 2D~lidar sensor of $360^\circ$ angular and \SI{6}{\meter} linear range that is downsampled via min-pooling to $N_L=30$ rays in \mbox{$\mathcal{L}_t = \{ {d}_i^t | 0 \leq i < N_L \}$}.
We include the local goal position $\boldsymbol{p}_g$ and human position~$\boldsymbol{p}_h$ in robot-centric polar coordinates.
The resulting state space vector is $s_t = \left(\boldsymbol{p}_g, \boldsymbol{p}_h, \mathcal{L}_t\right)$.
$\mathcal{L}_t$ and $(\boldsymbol{p}_g, \boldsymbol{p}_h)$ are processed in separate MLP feature extractors of 2 layers with dimensionality 64, before the two data streams are concatenated and processed by the TD3 actor and critic networks of size $128 \times 400 \times 300 \times \{1,2\}$.

\subsubsection{Action Space}
The policy outputs velocity control commands of linear and angular velocity as $a_t = (v, \omega)$ that directly drive the robot within a range of $v \in [0, 0.5] \u{\meter\per\second}$ and $\omega \in [-\pi, +\pi] \u{\radian\per\second}$.

\subsubsection{Reward}
For the navigation task, we employ the reward function of basal navigation objective such as goal-pursuance and collision avoidance as
\begin{equation}~
	\label{eq:base_reward}
	r^t = r^t_\textit{goal} + r^t_\textit{collision} + r^t_\textit{timeout} + r^t_\textit{loop} \text{.}
\end{equation}
The goal-pursuance is encoded in a sparse reward $r^t_\textit{goal} = c_\textit{goal}$ for arrival at the goal location such that $d_\textit{g} = |\boldsymbol{p}_g| \leq \SI{0.4}{\meter}$.
To encourage collision-free navigation, a sparse penalty $r^t_\textit{collision} = c_\textit{collision}$ is provided upon collision.
The sparse penalty $r^t_\textit{timeout} = c_\textit{timeout}$ penalizes non goal-oriented behavior after $\num{500}$ time steps.
The last term $r^t_\textit{loop}$ will be explained in Sec.~\ref{sec:ensemble_and_regularization}.

\subsubsection{Training Environment}
For training, we use the iGibson simulator~\cite{li_igibson_2022} that itself relies on the Pybullet physics engine~\cite{coumans_pybullet_2016}.
In an open space, we first randomly sample a start and goal location for the robot in a range of \SIrange{2}{10}{\meter}.
A single human is placed between the start and goal locations.
Subsequently, we sample a total of four cubic obstacles at random locations in the vicinity of the human and robot, avoiding the already occupied poses.
One episode is resembled by the rollout of one trajectory in a given environment configuration until one of the following three termination criteria is satisfied:
Timeout at more than $\num{500}$ time steps or $\SI{100}{\second}$, or successfully reaching the goal position within a threshold of $d_\textit{g} \leq \SI{0.4}{\meter}$.

\section{Our Approach}
This section introduces EnQuery with respect to the policy ensemble, the querying methodology, reward model training, and subsequently policy alignment.
Ultimately, we present a novel behavior explainability visualization.
\subsection{Ensemble Generation}
\label{sec:ensemble_and_regularization}
We extend the standard single-policy reinforcement learning architecture by introducing a set of $N_E$ policies $\mathcal{E} = \{\pi_i(s_t, a_t)| i \in [N_E]\}$, called the policy ensemble.
During training, each ensemble policy is interacting with its own environment instance, and storing the collected experiences into its own replay buffer of size $N_B$.
To achieve behavioral diversity across the ensemble, we use a regularization term to penalize similar outputs.
So as a core modification to achieve a diversity of behaviors for a given state $s_t$, we introduce the novel goal-modulated diversity regularization (GMDR) term
\begin{equation}
	\label{eq:diversity}
	\mathcal{L}^i_\textit{GMDR} = - \tilde{\kappa} \cdot \alpha_\textit{dist}(d_g) \cdot \sum_{j=0, j \neq  i}^{N_E} (a_i - a_j)^2 \text{,}
\end{equation}
which captures the difference between all pairwise combination of action outputs $\pi_i(s_t) = a_i$ of the ensemble members~$i$.
Here, the scaling factor $\tilde{\kappa} = \kappa/|A|^2$ is normalized by the dimension of the two-dimensional action space $A$.

A task-specific feature is the goal distance weighting term $\alpha_\textit{dist}(d_g) = m_\textit{dist} \cdot d_g + b_\textit{dist}$ that linearly decays the diversity loss with decreasing distance to the goal $d_g$.
The variables $m_\textit{dist}$ and $b_\textit{dist}$ normalize the term for the expectable distance range to the goal.
As a practical motivation, the closer the robot navigates to the goal, the fewer deviations from goal-directed navigation behavior are desired.
This helps the convergence of policy training from the goal-reaching perspective, while allowing for greater trajectory diversity when the goal is still far enough away.

In the TD3 architecture, the GMDR is simply added to the loss of the actors as
\begin{equation}
\mathcal{L}^i_{\text{actor}} = \mathcal{L}_{\text{actor}} + \mathcal{L}^i_\textit{GMDR} \text{.}
\end{equation}

We furthermore introduce the reward term $r^t_\textit{loop}$ as a countermeasure for undesired looping behavior that some policies of the ensemble would adapt as a result of the diversity regularization term, see Eq.~\ref{eq:base_reward}.
Essentially, the looping penalty checks for the self-intersection of the current trajectory, which is sparsely penalized with $r^t_\textit{loop} = c_\textit{loop}$.
Note that the criterion for self-intersection is applicable solely to trajectory segments that are more than four time steps old.
In all other cases, the sparse rewards are zero.
The reward function is identical for all agents in the ensemble, and explicitly not the source of ensemble diversity.

Before the ensemble policy is trained, a single policy $\pi_\text{raw}$ is trained without GMDR, but on the same reward (Eq.~\ref{eq:base_reward}) and task.
Subsequently, the replay buffers of the ensemble is initialized with the experiences of the RAW policy, which supports ensemble convergence and training success.
Also, the all policy members are initialized with the weights of $\pi_\text{raw}$.
Finally, the ensemble policies train for $T=\num{25}$k time steps, using the learning parameters denoted in Table~\ref{tab:params}.
Subsequently, the ensemble is ready to generate trajectory suggestions as queries for a single navigation task.

\begin{table}[!b]
	\centering
	\begin{tabular}{lcl}
		Notation & Value & Description \\
		\toprule
		$\gamma$ 	& 0.98	& RL discount value\\
		$T$ 		& 25e3 	& Training time steps\\
		$N_B$ 	& 1e6	& Replay buffer size\\
		$N_L$ 	& 30	& Pooled lidar ray number\\
		$c_\text{goal}$ 	& 20	& Sparse goal reward\\
		$c_\text{collision}$ 	& -1.2	& Sparse collision reward\\
		$c_\text{timeout}$ 	& -20.0	& Sparse timeout reward\\
		$c_\text{loop}$ 	& -2	& Sparse looping reward\\
		$N_E$ 	& 4 	& Ensemble member number\\
		$\kappa$ 	& 0.0625	& GMDR scaling factor\\
		$m_\textit{dist}$ 	& 1/8	& GMDR distance scaling slope\\
		$b_\textit{dist}$ 	& 1/4	& GMDR distance scaling intercept\\
		$k$ 	& 20	& BL query segment length\\
		$\lambda$ 	& 0.06	& Reward weighting factor\\
		\bottomrule
	\end{tabular}
	\caption{Notations and parameter settings. \label{tab:params}}
\end{table}

\subsection{Querying}
Our approach adopts pairwise A$\succ$B preference comparisons as a feedback modality.
Preferences $\succ$ are expressed over robot navigation trajectories $\tau = \{(s_0, a_0), (s_1, a_1), ..., (s_T, a_T)\}$, represented by state-action pairs.
We indicate $\tau_1 \succ \tau_2$ to indicate preference of trajectory $\tau_1$ over $\tau_2$.

Our ensemble of policies $\mathcal{E}$ generates $N_E$ trajectories for a given environment configuration.
From this, we randomly sample a trajectory pair.
All self-intersecting and collision-flawed trajectories are filtered.
Since we do not conduct a user study but focus on the methodology, we simulate human preferences and query an oracle that will always prefer the trajectory of higher minimum human distance $d_h = |\boldsymbol{p}_h|$.
The resulting preference dataset is denoted as $\mathcal{D}_\text{ens} = \{\tau_1^i \succ \tau_2^i | i \in [N_Q]\}$ after $N_Q$ oracle queries.

\subsection{Baseline Querying Approach}
We adopt the segment-based uniform querying approach of Christiano~\etal~\cite{christiano_deep_2017} as a baseline.
They achieve diversity not by an ensemble but via randomization of the environment, which translates to randomly generated start, goal, human, and obstacle positions for our environment.
A pool of trajectories is generated using the non-ensemble deterministic policy $\pi_\text{raw}$, from which we uniformly sample trajectory segments $\sigma$ with a length of $k = 20 \simeq \SI{4}{\second}$.
The preference is subsequently expressed over trajectory segments as $\sigma_1 \succ \sigma_2$, where $\sigma = \{(s_0, a_0), (s_1, a_1), ..., (s_{k-1}, a_{k-1})\}$ denotes a segment sampled from a trajectory $\tau$.
The resulting preference dataset is denoted with $\mathcal{D}_\text{seg} = \{\sigma_1^i \succ \sigma_2^i | i \in [N_Q]\}$.

\subsection{Reward Model}
To align the navigation policy, we first train a reward model $\hat{R}(s,a)$ from the pairwise preference dataset $\mathcal{D}$ based on the Bradley-Terry model~\cite{bradley_rank_1952}, where
\begin{equation}
	\hat{P}[\tau_1 \succ \tau_2] = \frac{1}{1 + \exp \left(R(\tau_2) - R(\tau_1)\right)}
\end{equation}
denotes the probability of a human preferring segment $\tau_1 \succ \tau_2$ with the cumulative return $R(\tau_i) = \sum_{(s, a) \in \tau_i} \hat{R}(s,a)$.
On that basis, a neural network is trained using a cross-entropy loss such that $\sum_{(s, a) \in \tau_i} \hat{R}(s,a) < \sum_{(s, a) \in \tau_j} \hat{R}(s,a) \text{ when } \tau_i \prec \tau_j$.
The reward model shares the network architecture with the critic and is trained for $\num{10}$ epochs using a learning rate of \num{1e-4}, after which the best-performing epoch model is chosen.
The output of the reward model is normalized to a distribution mean of zero with standard deviation one.

\subsection{Policy Alignment}
Preference-alignment of the navigation policy starts with a converged policy $\pi_\text{raw}$, as introduced above.
We take inspiration from the work of Cabi \etal~\cite{cabi_scaling_2020}, who recycle already collected data by updating the existing data buffer with the current reward model.
So with data efficiency in mind, we solely rely on the existing replay buffer data for alignment.
In other words, it is not necessary during alignment to further explore the environment. 
A subsequent batch-based policy update on the previous but reward-updated experiences aligns the policy.

To ensure that the alignd policy still obeys the basal navigation objectives defined by Eq.~\ref{eq:base_reward}, we balance between the preference reward model and the basal task reward for the updated reward
\begin{equation}
	r^*_t = \lambda \hat{R}(s_t,a_t) + (1-\lambda) r_t
\end{equation}
using the weighting factor $\lambda = \num{0.06}$ determined in preliminary experiments using a grid search.
During alignment of $\pi_\text{raw}$, we sample batches from the reward-updated replay buffer as usual and perform 10k policy updates each for one alignment epoch.
The models are tested for their navigation success rate after each epoch, where the best-performing epoch is chosen.

\subsection{Explainability Navigation Plot}
\label{sec:streamplot}
In autonomous robot navigation, the interpretability of reinforcement learning (RL) policies is essential.
Understanding and foreseeing the robot's actions is key to trust and acceptance, necessitating the development of tools that explain decision-making and ensure it meets human standards.

Recent explainability efforts for RL navigation policies target the reasoning pipeline and resulting behavior~\cite{he_explainable_2021, cruz_explainable_2023}. 
Other works project the learned Q-values into the scene~\cite{heuillet_explainability_2021}.
Yet, we found no visualization to give a complete picture of the behavior across the entire navigation scenario, an important tool to study the quality of navigational preference alignment.

We introduce a novel behavior explanation and visualization method for the navigation policy in a static environment, see Fig.~\ref{fig:c1}.
Based on the concept of a flow fields, it extracts and condenses the preferred navigation direction at all locations at once into a comprehensive birds-eye-view plot.
Firstly, we discretize the environment into a 2D grid of \SI{0.25}{\meter} resolution and place the robot at the center of each traverseable cell oriented towards the goal.
While keeping the forward velocity at zero, we solely execute the angular velocity command. 
The robot turns and settles like a compass needle into a certain direction.
Whenever the settling results in an oscillation around one direction, we take the mean of the oscillation range.
Subsequently, the obtained directional driving preference is recorded together with the magnitude of the corresponding forward velocity.
We obtain a matrix of 2D velocity vectors as in a flow field that are visualized using a stream plot.
The stream plot sketches the driving behavior across the entire scene at once.

In a second step, we reactivate the forward velocity and roll the trajectories out from each grid cell.
Subsequently, the number of traversals through each grid cell is counted to to create a heat map, which is visualized behind the stream plot.
This completes the picture especially at locations of ending streamlines, whenever the streamline density is too high.
For plotting, we use the python library matplotlib~\cite{hunter_matplotlib_2007}.
Note that the plotting scheme assumes that the robot is the only entity moving within the scene.

\section{Experimental Evaluation}
\label{sec:exp}
Our experiments investigate the ensemble diversity and success rate with respect to the regularization term Eq.~\ref{eq:diversity} in qualitative and quantitative measures, the query and reward learning process in comparison to a well established baseline approach, and finally the preference-alignment of the resulting navigation policy.
 
\subsection{Ensemble}
\begin{figure}[t]
	\centering
	\includegraphics[width=0.99\linewidth]{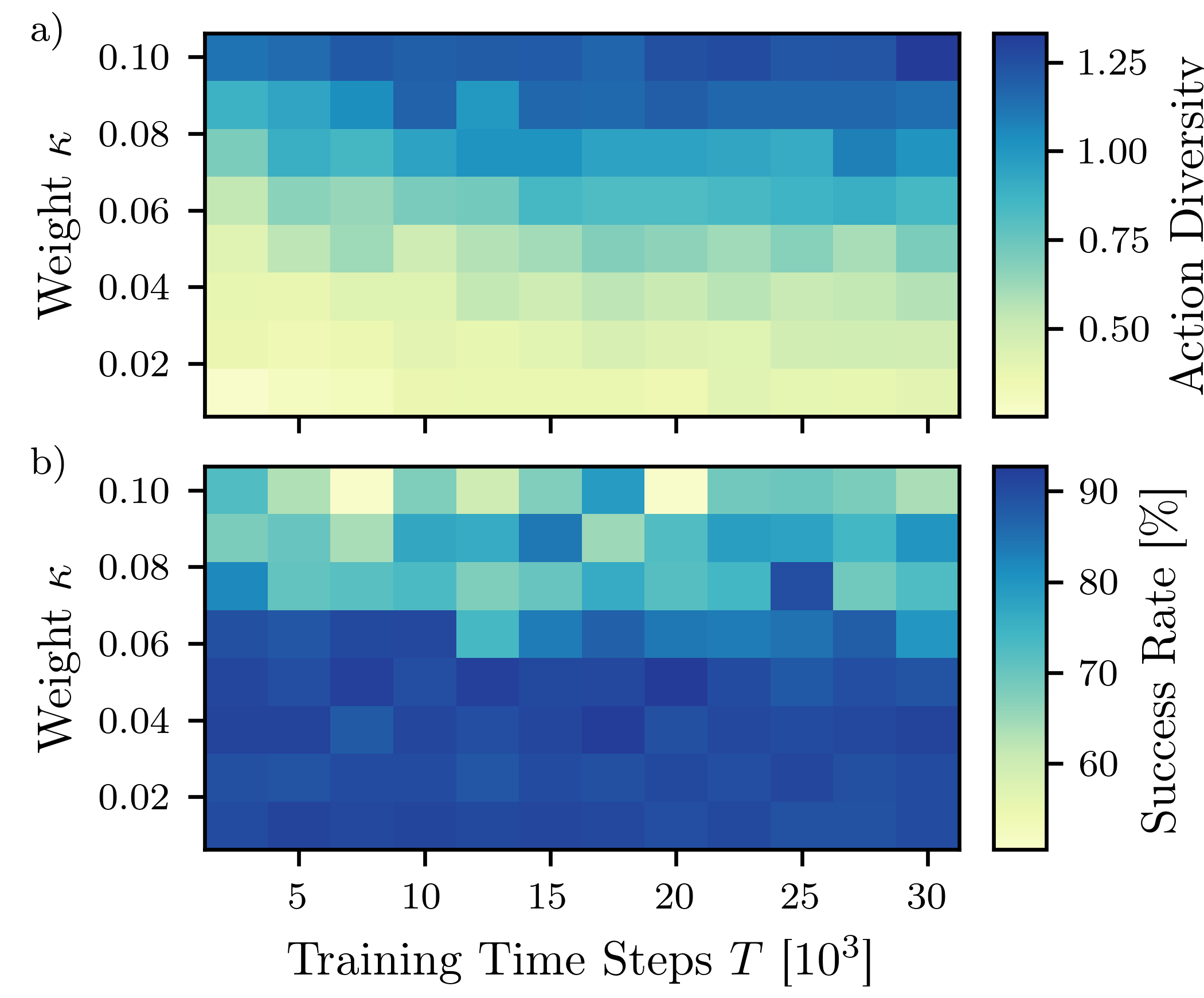}
	\caption{
		\textbf{a)} Diversity of actions over the ensemble and \textbf{b)} success rate on the navigation task in dependence of the total training time steps $T$ and the weighting factor $\kappa$ of the regularization term.
		The action diversity grows with the weight $\kappa$ of the regularization term, while the success rate decreases rapidly for $\kappa > \num{0.07}$.
	}
	\label{fig:a1}
\end{figure}
\begin{figure}[b]
	\centering
	\includegraphics[width=0.99\linewidth]{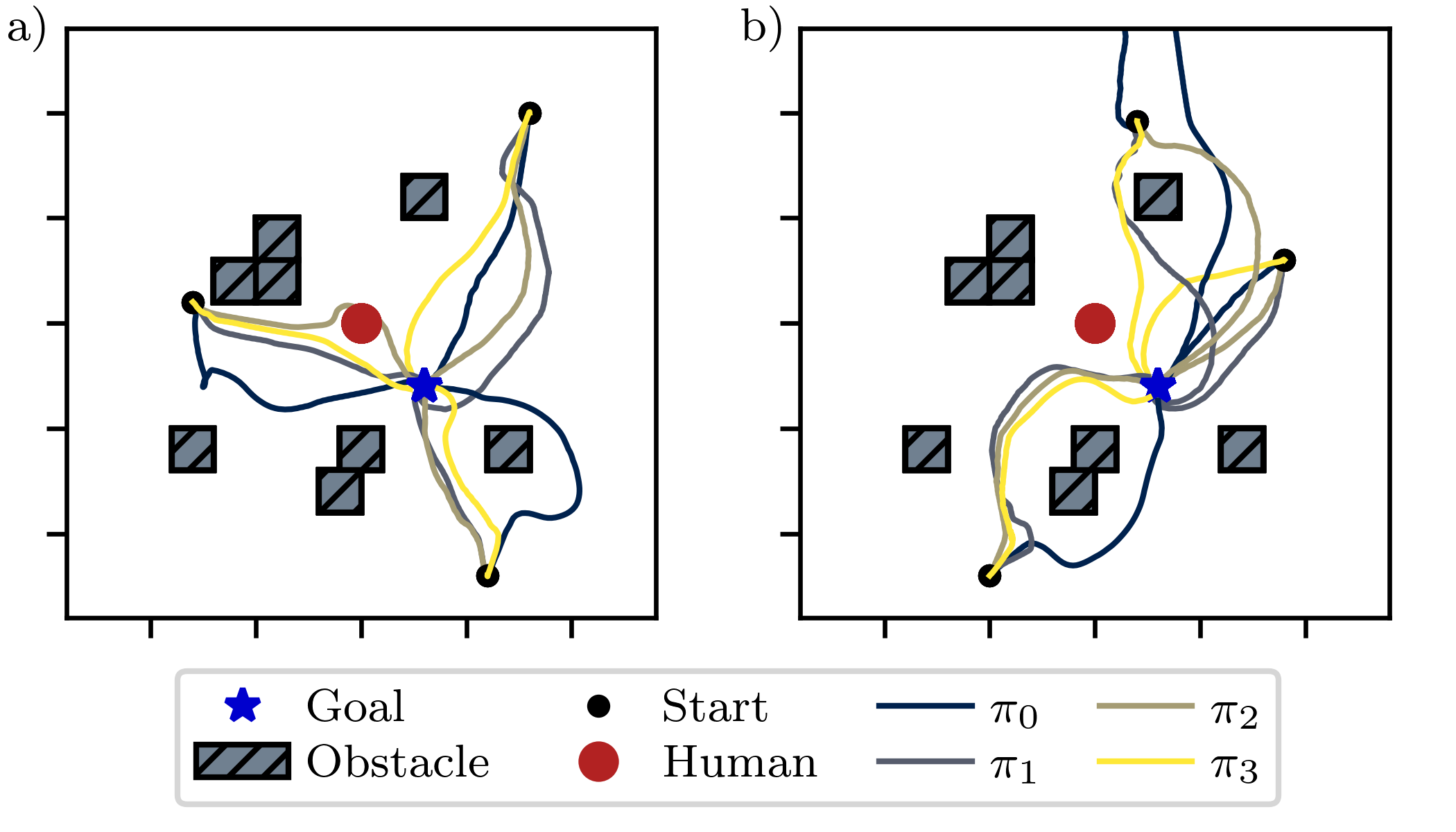}
	\caption{
		Trajectories of the ensemble policies $\pi_i$ for a given obstacle configuration and randomized start position.
		Each plot \textbf{a)} and \textbf{b)} shows three individual start positions.
		A distinct diversity of the trajectory pathways can be observed.
	}
	\label{fig:a3}
\end{figure}
\subsubsection{Quantitative}
The query ensemble is based on a set of $N_E=4$ policies that obey the GMDR.
First, we evaluate the influence of the $\kappa$-scaled GMDR on the learning behavior and diversity of the ensemble.
In dependence of the total training time steps and the GMDR's scaling factor~$\kappa$, the raw action diversity $\sum_{j=0, j \neq  i}^{N_E} (a_i - a_j)^2$ is computed for 1,000 randomly sampled states from the replay buffer, see Fig.~\ref{fig:a1}a.
Also, the success rate averaged of 100 trajectories and all ensemble policies visualized, see Fig.~\ref{fig:a1}b.
Generally, an increasing action diversity can be observed with growing scale of the regularization term, while the success rate decreases rapidly for $\kappa > \num{0.07}$.
Furthermore, the diversity grows with increasing training time steps, without an obvious decrease of the success rate.

Based on the grid search, we settled for an optimal configuration of $\kappa = 0.0625$ and $T=$ 25k training time steps.
Here, the ensemble achieves a success rate of \SI{91}{\percent} at an action diversity average of $\kappa = \num{0.7}$.
To put this in contrast, the non-ensemble agent $\pi_\text{raw}$ achieves a success rate of \SI{94}{\percent}.

In preliminary experiments have also experimented with an regularization penalty that was not goal distance-scaled and $\alpha_\textit{dist} = 1$, resulting in lower success rates for similar action diversity. 
For optimizing our task setup, we found the goal-modulation beneficial.

\subsubsection{Qualitative}
We visualize the trajectories of all ensemble policies $\pi_i$ in Fig.~\ref{fig:a3} for a given environment configuration of eight obstacles.
The trajectory shapes vary due to the enforced output diversity, as expected.
The diversity spans from avoiding obstacles on the other side (compare $\pi_0$), to keeping different distances to the human in the vicinity of the robot.
If straight-line navigation to the goal is possible, one policy usually take the shortest route while the others meander in more curvy and longer trajectories.

\subsection{Reward Model}
\label{sec:results_rm}
\begin{figure}[t]
	\centering
	\includegraphics[width=0.99\linewidth]{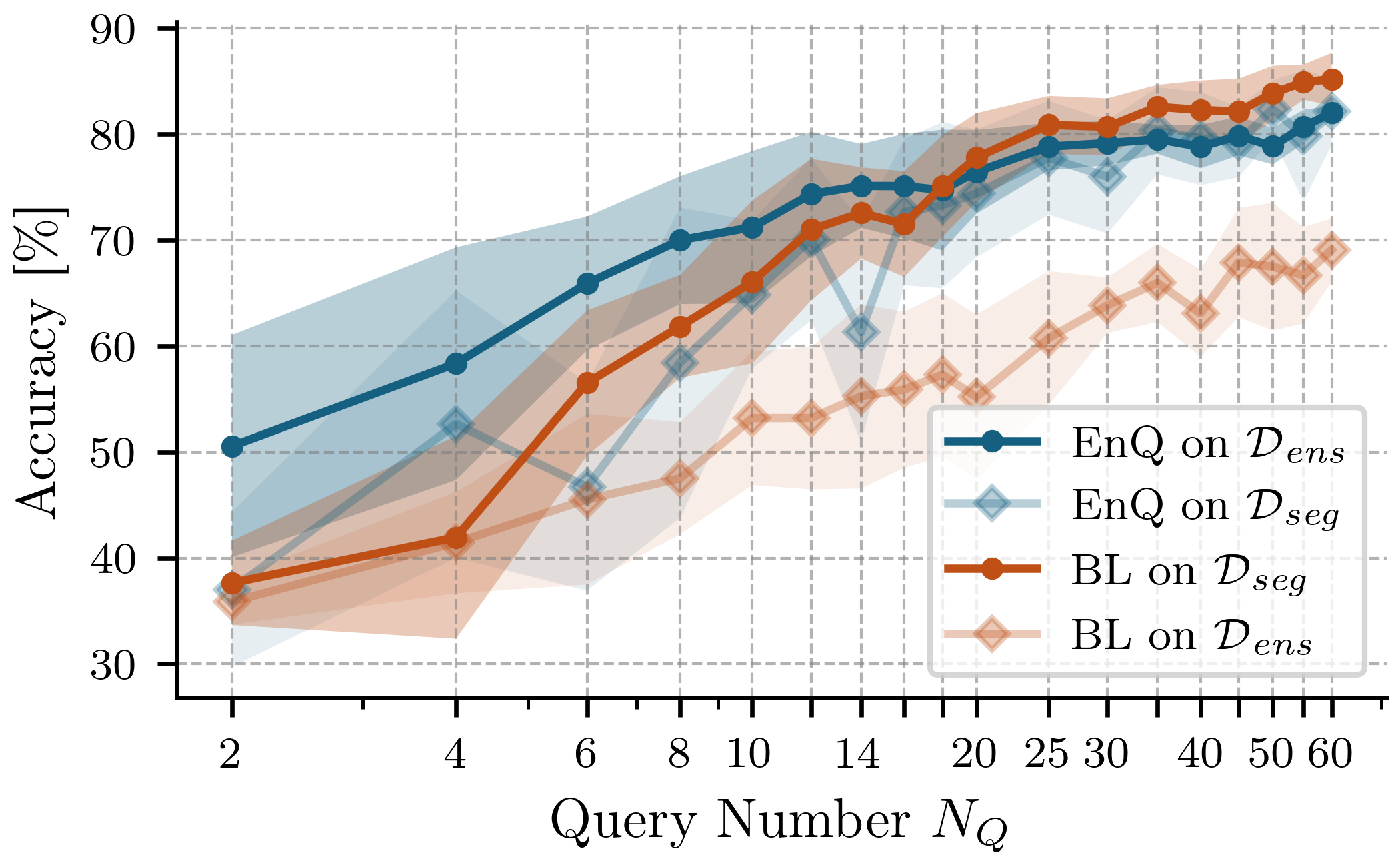}
	\caption{
		Reward model test accuracy for our EnQ approach and the baseline of segment-based uniform sampling~\cite{christiano_deep_2017} over different query numbers on their native dataset (e.g., EnQ on $\mathcal{D}_\textit{ens}$) and in cross validation (e.g., EnQ on $\mathcal{D}_\textit{seg}$)
		The process of querying, reward model training, and testing has been repeated ten times, for which mean and standard deviation are shown.
		We outperform the baseline with a higher test accuracy thus information gain for low query numbers, enabling a faster learning curve time-critical learning scenarios. 
	}
	\label{fig:b2}
\end{figure}
\begin{figure*}[t]
	\centering
	\includegraphics[width=1.0\linewidth]{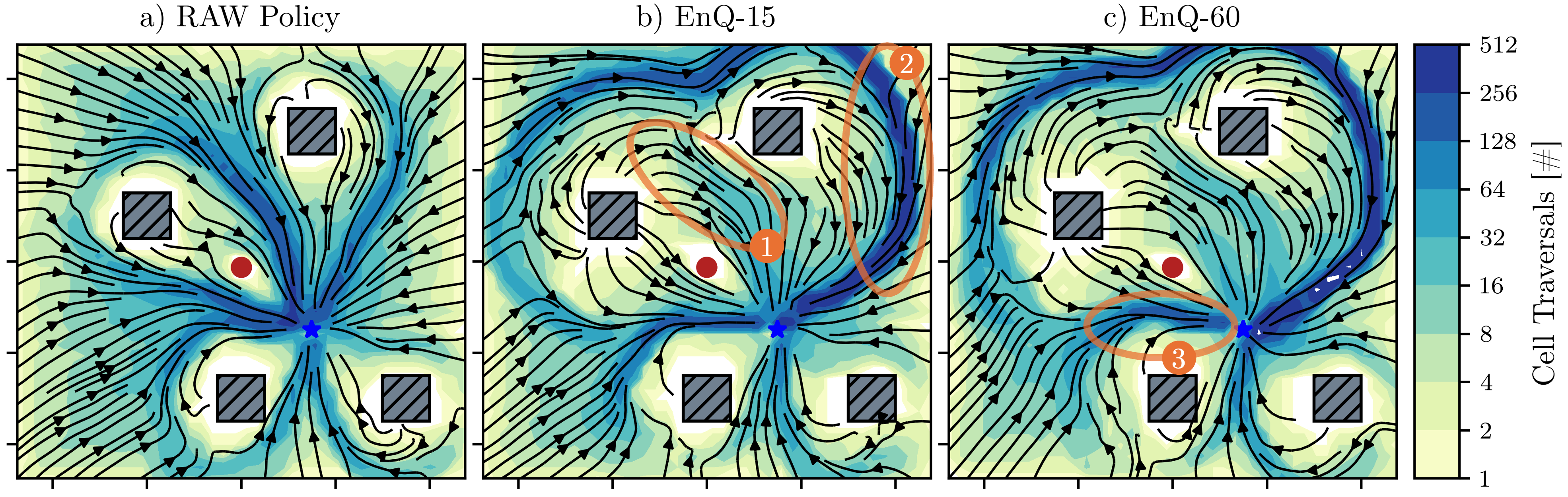}
	\caption{
		Driving behavior for a given scene visualized by our novel explainability navigation plot, compare Sec.~\ref{sec:streamplot} for \textbf{a)} the raw policy~$\pi_\text{raw}$, \textbf{b)} the preference-aligned policy EnQ for $N_Q = 15$ queries, and \textbf{c)} for $N_Q = 60$ queries.
		The trajectory flow can be derived from any start position in the given scene to the goal (blue star), while circumnavigating the human (red dot).
		Regions of interest (ROI) are indicated in orange.
		Under the raw policy, mostly goal-directed and collision-avoiding navigation behavior can be observed.
		For the aligned policies, a pronounced shift away from the human at the cost of longer trajectories, e.g., on the far side of the top right obstacle appears (ROI 2).
		At the same time, traversal wise the area around the human is thinned out (ROI 1), as indicated by the underlying traversal map.
		EnQ-60 traverses closer to the human in the direct vicinity (ROI 3).
	}
	\label{fig:c1}
\end{figure*}
\begin{figure}[t]
	\centering
	\includegraphics[width=1.0\linewidth]{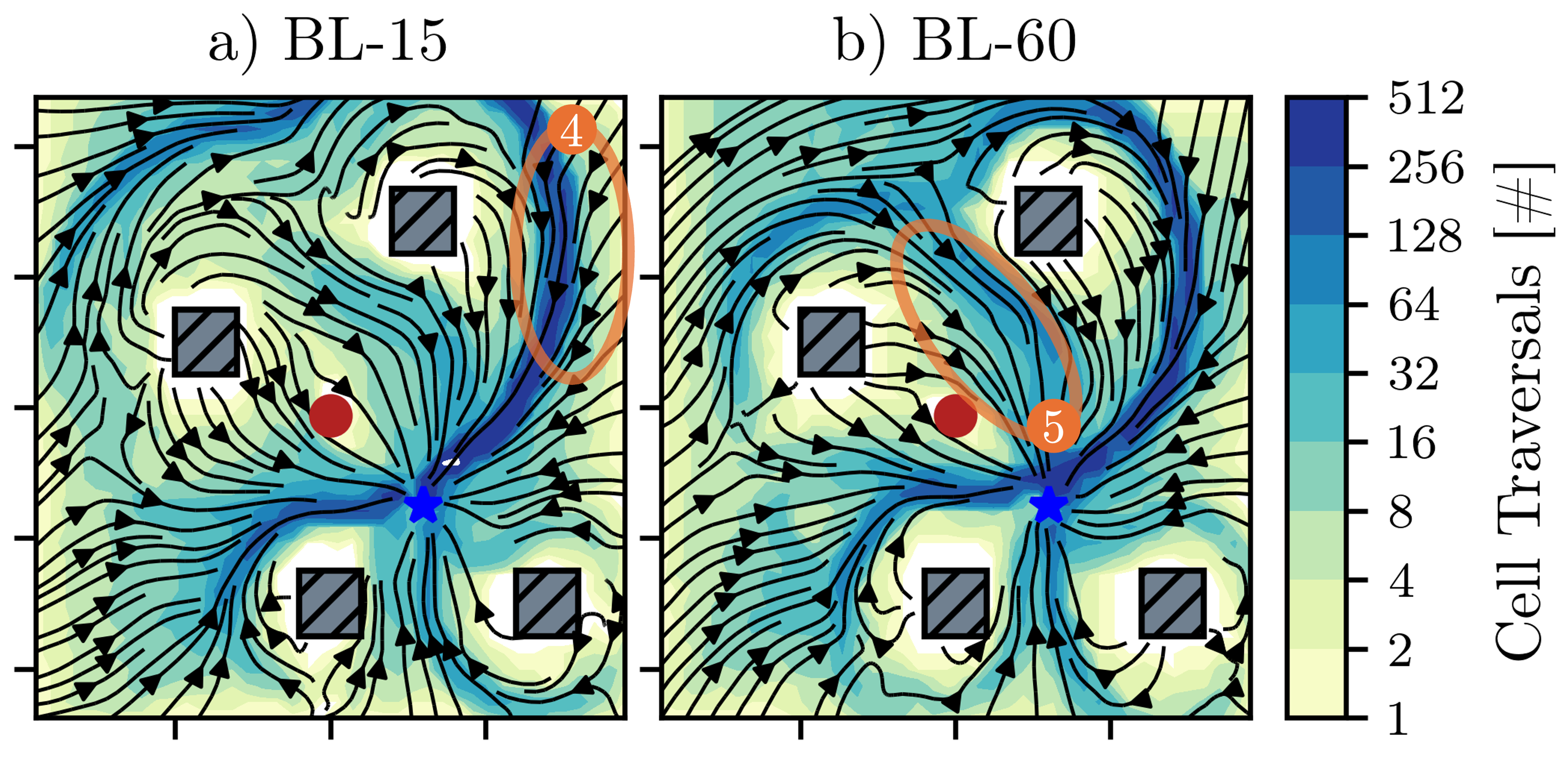}
	\caption{
		Driving behavior for a given scene visualized by our novel explainability navigation plot, compare Sec.~\ref{sec:streamplot} for the baseline segmentation-based approach with \textbf{a)} $N_Q = 15$ and \textbf{b)} $N_Q = 60$ queries.
		Comparing to EnQ (see Fig.\ref{fig:c1}), the developed outer traversal corridor on the right falls closer to the human (ROI 4).
		Furthermore on BL-60, the corridor directly above the human (ROI 5) is more traversed more often as compared to EnQ-60, indicating less distance-keeping from the human.
	}
	\label{fig:c1_bl}
\end{figure}
We analyze the information gain for our ensemble query method (EnQ) against the baseline (BL) of segment-based uniform sampling by Christiano \etal~\cite{christiano_deep_2017}.
Our measure for the information gain is the prediction accuracy of the reward model on a test dataset in dependence of the number of queries $N_Q$.
Specifically, we query the oracle that prefers trajectories with higher human distance for a total of $N_Q$ times to generate a preference dataset using both our ensemble queries ($\mathcal{D}_\textit{ens}$) and the segment baseline ($\mathcal{D}_\textit{seg}$).
The accuracies are tested on test similar-sized splits of both the ensemble and segment datasets, respectively, providing both a normal and a cross validation.
As can be seen in Fig.~\ref{fig:b2}, we outperform the segment-based baseline approach for lower query numbers.
For higher query numbers $N_Q \geq 18$, the baseline achieves a higher test accuracy.
Notably, the segment-based reward model does not generalize well to the ensemble data, while the ensemble-based reward model generalized well to the segment dataset $\mathcal{D}_\textit{seg}$. 
We can conclude that EnQ provides an advantage in information gain feedback processes that need to be query efficient.

\subsection{Policy Alignment}
\begin{figure}[t]
	\centering
	\includegraphics[width=1.0\linewidth]{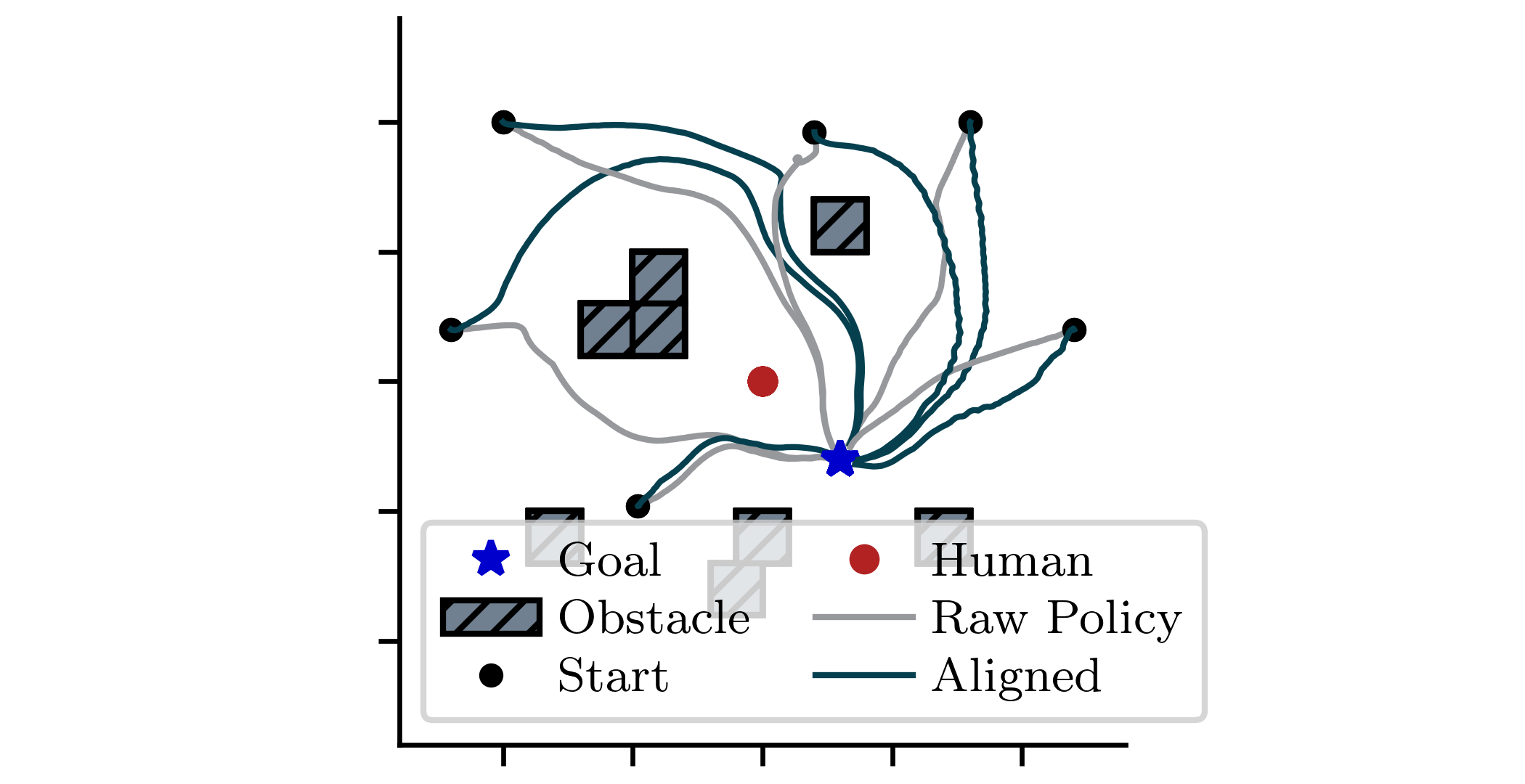}
	\caption{
		Trajectories of the raw and aligned policies $\pi_\text{raw}$ and $\pi_\text{aligned}$ for a given obstacle configuration and randomized start position.
		While the raw policy exhibits mostly goal-directed navigation behavior, $\pi_\text{aligned}$ reflects the preference of maximum human distance keeping on a majority of the trajectories.
	}
	\label{fig:c2}
\end{figure}
The following experiments target the final navigation performance of the preference-aligned policy $\pi_\text{aligned}$ that should keep a high distance to the human.
The aligned policies will be denoted as EnQ for our ensemble-based approach EnQuery and BL for the baseline.
Those model names are complemented by the number of queries $N_Q$ used to align the model.
We chose $N_Q = 15$ as a low query and $N_Q = 60$ as a high query number, where the EnQ reward model outperforms the BL in the low query, and vice versa for the high query regime, compare Sec.\ref{sec:results_rm}.
\subsubsection{Quantitative}
\setlength\lightrulewidth{0.1ex}
\begin{table}[b!]
	\centering
	\begin{tabular}{lccccc}
		& ${\min(d_h)}$ [m] & SR$\uparrow$ & CR$\downarrow$ & TR$\downarrow$ [\%] & $p$\\
		\toprule
		\textbf{EnQ-15} & \textbf{3.2} & 89.8 & \textbf{3.4} & 6.8 & -\\
		EnQ-60          & 2.1 & 92.4 & 3.9 & 3.7 & ** \\
		BL-15           & 2.5 & \textbf{93.9} & 3.7 & \textbf{2.4} & *** \\
		BL-60           & 2.7 & 86.8 & 3.6 & 9.6 & *** \\
		RAW             & 1.2 & 94 & 6 & 0 & ***\\
		\bottomrule
	\end{tabular}
	\caption{
		Quantitative analysis of the performance and minimum human distance $\min(d_h)$ averaged over 10 alignment runs and 100 trajectory rollouts with identical scene setups.
		EnQ-15 denotes the ensemble-query aligned model of $N_Q = 15$, EnQ-60 of $N_Q = 60$ queries. 
		Analogously, BL-15 and BL-60 denote the segment-based baseline querying approach with uniform sampling~\cite{christiano_deep_2017}.
		RAW denotes the non-aligned initial policy, averaged over 100 trajectory rollouts.
		SR, CR, and TR denote the success, collision, and timeout rate, respectively.
		Statistical significance of $\min(d_h)$ for dissimilar distribution means against EnQ-15 is denoted in last column $p$, where * for $p\leq0.05$, ** for $p\leq0.01$, *** for $p\leq0.001$.
		EnQ-15 as our flagship approach exhibits the best alignment in terms of the preference metric, while running into more timeouts due to possibly longer trajectories.
	}
	\label{tab:ablation}
\end{table}
The ablation study presented in Table~\ref{tab:ablation} quantitatively evaluates the performance of different configurations of a navigation system across several metrics: success rate (SR), collision rate (CR), timeout rate (TR), and minimum human distance (\(\min(d_h)\)) as our metrics.
The results were averaged over 10 aligned policies with different query samples and 100 trajectory rollouts with identical scene setups.
For the preference metric \(\min(d_h)\), we test for statistical significance of EnQ-15 exhibiting higher $\min(d_h)$ as compared to the other configurations using a student's t-test, as denoted in the last column of Table~\ref{tab:ablation}.
We compare the aligned models against the RAW policy $\pi_\text{raw}$, specifically in terms of distance to the human in alignment.
The ensemble-query aligned model with only 15 queries denoted as EnQ-15 significantly outperforms all other configurations in terms of the highest minimum human distance.
EnQ-60 exhibits a higher success rate compared to EnQ-15, but at the cost of weaker distance keeping.
The baseline policy BL improves with respect to $\min(d_h)$ as the number of queries increases from 15 to 60.
In this regime, our approach EnQ-15 achieves the lowest collision rate, at the cost of more timeouts as compared to BL-15.
Logically, timeouts become more likely as the robot drives longer, more human-distant trajectories.
BL-15 exhibits the highest success rate, but at a lower preference metric compared to EnQ-15.
To conclude, our approach outperforms the baseline with respect to the preference metric for the low-query regimes, reflecting the results of improved reward model below $N_Q = 18$ queries, see Fig.~\ref{fig:b2}.

\subsubsection{Qualitative}
We compare the aligned policy EnQ-15 against $\pi_\text{raw}$ (RAW) and EnQ-60 qualitatively for a given navigation scenarios, see Fig.~\ref{fig:c1} with indicated regions of interest (ROI).
With regards to the baseline reward, both policies exhibit obstacle-avoidance and goal-pursuance behavior.
For the aligned policy, however, the driving patterns bend away from the human, as compared to goal-directed driving directions with $\pi_\text{raw}$.
With respect to the human this manifests, e.g., in an accumulation of trajectories on the far side around the top-right obstacle (ROI 2), allowing for more distance to the human.
Directly around the human, a thinning of robot's passages can be observed (ROI 1), with less trajectories approaching from the top-left of the plot alongside the human.
Furthermore, a noticeable outward bend of trajectories around the human (ROI 1) arises for the aligned agent.
Subtle but noticeable, EnQ-60 traverses closer to the human in the direct vicinity (ROI 3).

Comparing against the baseline approach BL-15 and BL-60 in Fig.~\ref{fig:c1_bl}, two findings strike.
The developed outer traversal corridor around the top-right obstacle falls closer to the human~(ROI 4) as compared to EnQ.
Furthermore on BL-60, the corridor directly above the human (ROI 5) is traversed more often as compared to EnQ-60 and BL-15, indicating less distance-keeping from the human.

A direct comparison of trajectories between $\pi_\text{aligned}$ and $\pi_\text{raw}$ is visualized in Fig.~\ref{fig:c2}.
Here, a similar picture manifests with the aligned trajectories traversing at a higher human distance, as compared to the goal-directed trajectories of the raw policy.

\section{Conclusion}
\label{sec:conclusion}
Our paper introduces EnQuery, a novel ensemble-based query method for diverse behavior suggestions in reinforcement learning from human feedback (RLHF) with deterministic policies.
We apply EnQuery to the field of robot navigation where the robot operates in the vicinity of humans who have specific preferences regarding the robot's navigation style. 
Using our output diversity regularization during training, the ensemble generates diverse trajectories which can be used to query preferences for any given navigation task. 
Importantly, generated queries maintain consistent reference points, such as the starting and goal position and use the same environment setup which improves the re-test reliability and thus the information extracted from the preference pairs.
The experiments show a superior information gain for low query numbers compared to a widely used baseline querying approach.
We then successfully demonstrate the data-efficient preference alignment of a navigation policy, by recycling the collected experiences data.
Finally, our novel method for visualizing navigation policy behavior comprehensively illustrates the alignment result. 

Looking ahead, we plan to conduct a comprehensive user study to investigate the user convenience and test reliability of our generative ensemble policy queries.
EnQuery aligns with our broader vision of developing more intuitive, efficient, and human-centric approaches to customize robotic navigation behaviors.

\bibliographystyle{IEEEtran}
\bibliography{bib_ensemble}
\end{document}